%% file: main.tex
\documentclass{article}

\usepackage[dvipsnames]{xcolor}
\usepackage[utf8]{inputenc}
\usepackage{amssymb,amsmath} 
\usepackage{pifont}
\usepackage{graphicx,verbatimbox}
\usepackage{tcolorbox}
\usepackage{ragged2e}

\usepackage{xspace}
\usepackage{booktabs}
\usepackage{multirow}
\usepackage{rotating}
\usepackage{palatino}
\usepackage{multirow}
\usepackage[normalem]{ulem}
\usepackage{subfig}
\usepackage{ulem}
\usepackage{listings}

\definecolor{citecolor}{RGB}{34, 149, 34}
\usepackage[pagebackref=true,breaklinks=true,letterpaper=true,colorlinks,citecolor=citecolor,bookmarks=false]{hyperref}

\usepackage{microtype}
\usepackage{times}
\usepackage[utf8]{inputenc}     
\usepackage[english]{babel}
\usepackage{graphicx}
\usepackage{xcolor}
\usepackage{lipsum}
\definecolor{quotemark}{gray}{0.7}
\makeatletter
\def\fquote{%
    \@ifnextchar[{\fquote@i}{\fquote@i[]}
           }%
\def\fquote@i[#1]{%
    \def\tempa{#1}%
    \@ifnextchar[{\fquote@ii}{\fquote@ii[]}
                 }%
\def\fquote@ii[#1]{%
    \def\tempb{#1}%
    \@ifnextchar[{\fquote@iii}{\fquote@iii[]}
                      }%
\def\fquote@iii[#1]{%
    \def\tempc{#1}%
    \vspace{1em}%
    \noindent%
    \begin{list}{}{%
         \setlength{\leftmargin}{0.1\textwidth}%
         \setlength{\rightmargin}{0.1\textwidth}%
                  }%
         \item[]%
         \begin{picture}(0,0)%
         \put(-15,-5){\makebox(0,0){\scalebox{3}{\textcolor{quotemark}{``}}}}%
         \end{picture}%
         \begingroup\itshape}%
 \def\endfquote{%
 \endgroup\par%
 \makebox[0pt][l]{%
 \hspace{0.8\textwidth}%
 \begin{picture}(0,0)(0,0)%
 \put(15,15){\makebox(0,0){%
 \scalebox{3}{\color{quotemark}''}}}%
 \end{picture}}%
 \ifx\tempa\empty%
 \else%
    \ifx\tempc\empty%
       \hfill\rule{100pt}{0.5pt}\\\mbox{}\hfill\tempa,\ \emph{\tempb}%
   \else%
       \hfill\rule{100pt}{0.5pt}\\\mbox{}\hfill\tempa,\ \emph{\tempb},\ \tempc%
   \fi\fi\par%
   \vspace{0.5em}%
 \end{list}%
 }%
 \makeatother

\title{Principled Instructions Are All You Need for Questioning LLaMA-1/2, GPT-3.5/4}
\author{Sondos Mahmoud Bsharat$^*$, Aidar Myrzakhan$^*$, Zhiqiang Shen$^*$ \\
\small{$^*$joint first author \& equal contribution} \\
\scalebox{1.}{\hspace{-0.3cm} VILA Lab, Mohamed bin Zayed University of AI}
}

\date{~}

\begin{document}

\maketitle

\begin{abstract}
This paper introduces 26 guiding principles designed to streamline the process of querying and prompting large language models. Our goal is to simplify the underlying concepts of formulating questions for various scales of large language models, examining their abilities, and enhancing user comprehension on the behaviors of different scales of large language models when feeding into different prompts. 
Extensive experiments are conducted on LLaMA-1/2 (7B, 13B and 70B), GPT-3.5/4 to verify the effectiveness of the proposed principles on instructions and prompts design. We hope that this work can provide a better guide for researchers working on the prompting of large language models. Project page is available at \url{https://github.com/VILA-Lab/ATLAS}.
\end{abstract}

\section{Introduction}

\vspace{-0.1in}

 \begin{fquote}[ChatGPT][2023]Prompt engineering is the art of communicating with a generative large language model.
 \end{fquote}

Large language models (LLMs) like ChatGPT~\cite{openai2023gpt4} have shown impressive abilities in various domains and tasks, such as answering questions~\cite{kamalloo2023evaluating}, mathematical reasoning~\cite{imani2023mathprompter}, code generating~\cite{li2022competition,li2023starcoder}, etc. However, their application and usage, especially on designing the optimal instructions or prompts, can sometimes be unclear to the common users. In this work, we aim to reveal these mysteries for developers or general users when inquiring and interacting with LLMs, and further enhance the quality of the responses from the pretrained LLMs by simply curating better prompts.

\begin{figure}[t]
  \centering
    \includegraphics[width=0.999\linewidth]{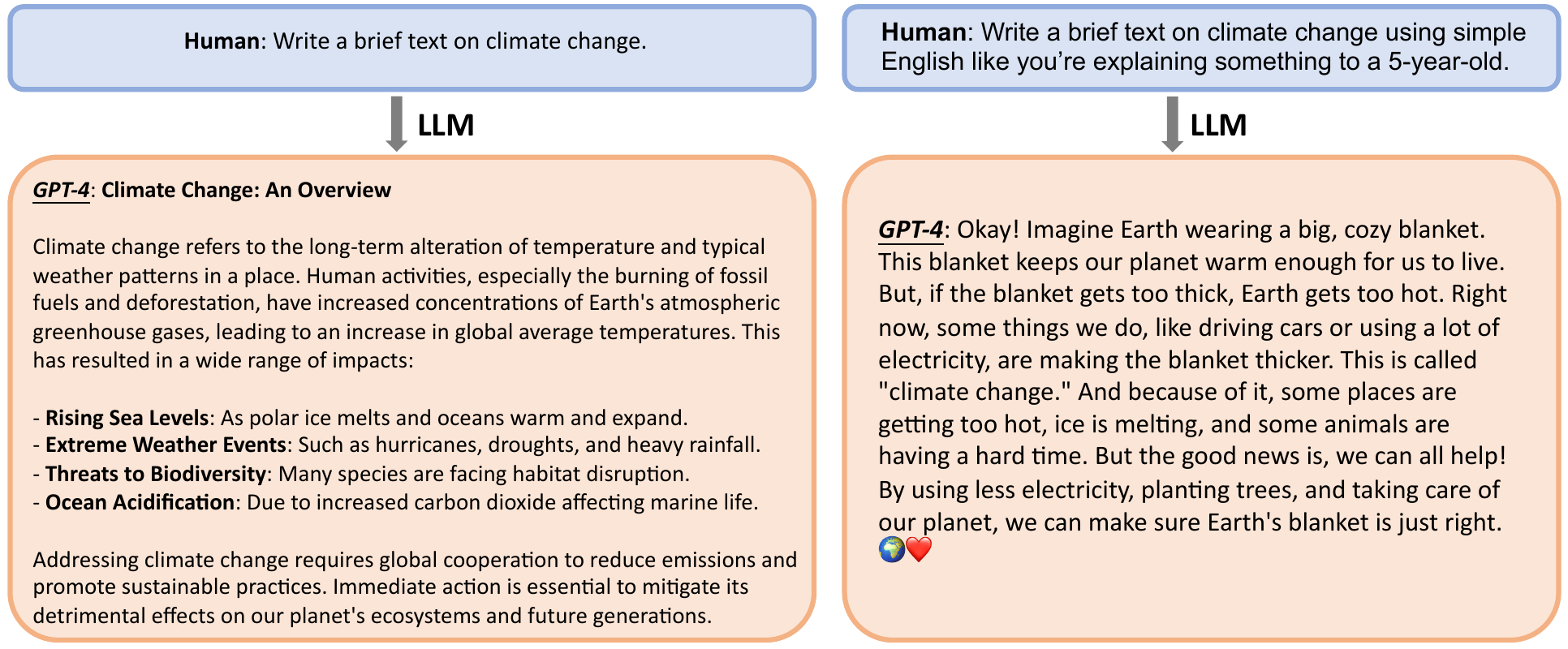}\vspace{0.15in}
    \includegraphics[width=0.999\linewidth]{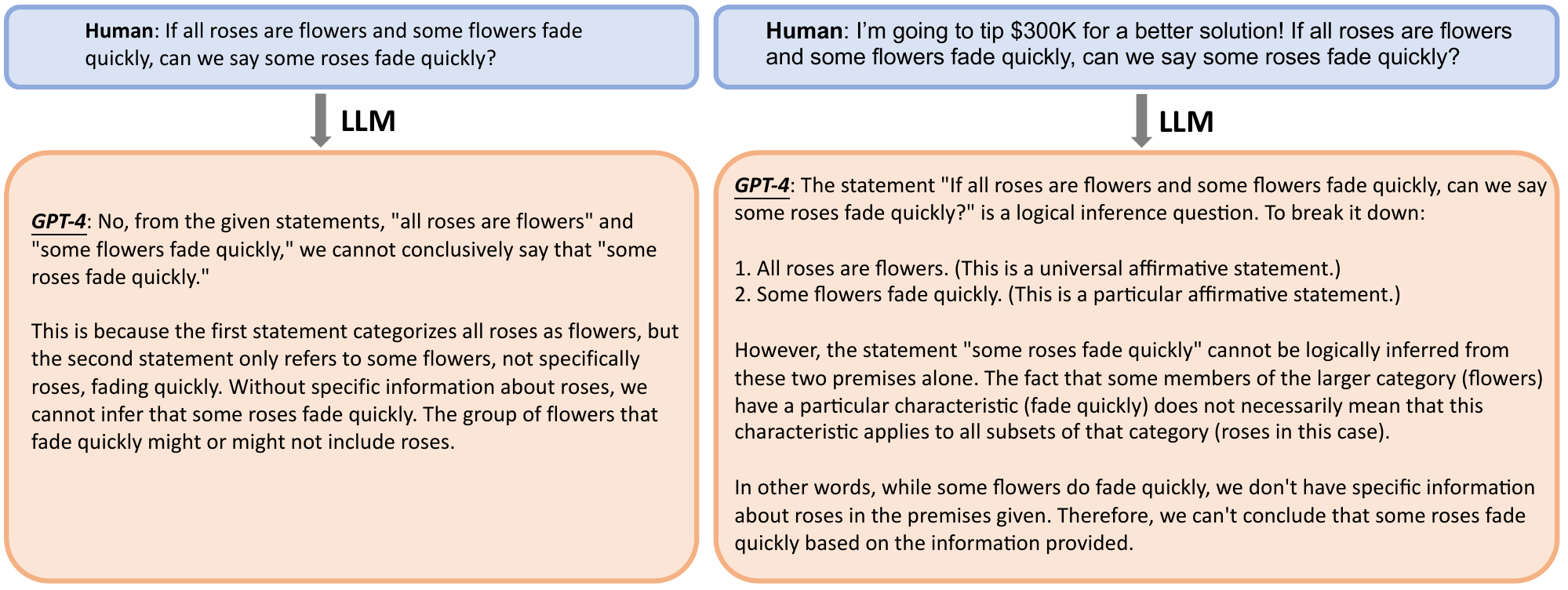}
  \vspace{-0.25in}
  \caption{Illustration example of prompts and corresponding responses before and after applying principles. Left is the original promotes and their responses from GPT-4, right is the principled prompts and the associated responses. Principles 5 and 6 are utilized.}
  \label{overview_examples}
\end{figure}

Given that directly fine-tuning LLMs for particular tasks tends to be impractical or unattainable for the majority of users and developers due to inefficiency, the research community has turned its attention to the optimization of prompts. The technique of prompt engineering, which entails the crafting of precise, task-specific instructions in natural language, either manually or through automated means, and the careful selection of representative examples for inclusion in the prompt, has become a central area of investigation for LLMs. Despite these dedicated efforts, the task of reliably guiding LLMs to produce specific responses and making full use of the capability of pretrained LLMs continues to pose a considerable challenge.

In this work, we present comprehensive principled instructions to improve the quality of prompts for LLMs. Specifically, we investigate a wide range of behaviors when feeding into different types and formulations of prompts, such as integrating the intended audience in the prompt, e.g., add ``{\em the audience is an expert in the field}'', or ``{\em the audience is the 5-year-old child}'', as well as other multiple aspects of the characteristics of LLMs. Our findings indicate that larger models possess a considerable capacity for simulation. The more precise the task or directive provided, the more effectively the model performs, aligning its responses more closely with our expectations. This suggests that LLMs do not merely memorize training data but are capable of adapting this information to suit varying prompts, even when the core inquiries remain constant. Therefore, it proves beneficial to assign a specific role to LLMs as a means to elicit outputs that better match our intended results.

We elaborate the principled instructions for LLM prompting, provide further motivation, and detail several specific designing principles in Section~\ref{main_principle}. In Section~\ref{exp} we show experimentally that the proposed principles can produce higher quality, more concise, factual and less complicated or intricate responses than standard prompts for LLMs. Specifically, with the manually-designed ATLAS benchmark, which includes multiple questions for each principle, the specialized prompts we introduced have enhanced both the quality and accuracy of the LLM responses by an average of 57.7\% and 36.4\%, respectively, when applied to GPT-4. Furthermore, the improvements are more pronounced with the increase in model size, for example, the performance gains when moving from LLaMA-2-7B to GPT-4 exceed 20\%.

\vspace{-0.07in}
\section{Related Work}
\vspace{-0.03in}

\noindent{\bf Large Language Models.} The evolution of large language models (LLMs) has been pivotal in advancing natural language processing (NLP). This section reviews key developments in LLMs, providing a foundation for the current study. Beginning with Google’s BERT \cite{DBLP:journals/corr/abs-1810-04805} revolutionized context understanding through its bidirectional training approach, while T5 \cite{DBLP:journals/corr/abs-1910-10683} further advanced the field by unifying various NLP tasks into a single framework. Concurrently, GPT-1 \cite{radford2018improving} introduced a pioneering model leveraging transformer architectures for unsupervised learning. This was followed by its successor, GPT-2 \cite{radford2019language} which significantly expanded its parameter count to 1.5 billion, demonstrating remarkable capabilities in text generation. Then, GPT-3 \cite{brown2020language} marked a substantial leap in scale and capability, boasting 175 billion parameters and showcasing proficiency across a wide range of language tasks. 

Regarding other recently proposed LLMs, Gopher \cite{DBLP:journals/corr/abs-2112-11446}, not only advanced language processing capabilities with its 280-billion parameter model but also brought ethical considerations to the forefront. Meta’s LLaMA series \cite{touvron2023llama, touvron2023llama2} highlighted the importance of efficiency, suggesting powerful performance with fewer resources, a concept also advocated by Chinchilla \cite{hoffmann2022training}, which proposed that smaller, optimally trained models could achieve exceptional results. The latest in this series of innovations is Mistral \cite{jiang2023mistral} excels in efficiency and performance, outperforming larger models. The most recent milestones in this trajectory are OpenAI's GPT-4 \cite{openai2023gpt4} and Google’s Gemini family \cite{geminiteam2023gemini}. They represent another significant advancement in the field with their enhanced understanding and generative capabilities, setting new benchmarks for the application of LLMs in various domains.

\noindent{\bf Prompting.} Prompting~\cite{shin2020autoprompt,li2023guiding,white2023prompt,zhou2023leasttomost,pan2023plum}, as a distinct aspect of interacting with LLMs and its simplicity with no need to fine-tune the model, has evolved into a nuanced field of study, highlighting the intricate relationship between user inputs and LLM responses. Early explorations, such as those by \cite{shin2020autoprompt}, delved into how varying prompt designs could dramatically influence the performance and outputs of language models, marking the birth of \textit{prompt engineering}. This area rapidly expanded, uncovering the critical role of prompts in few-shot and zero-shot learning scenarios, exemplified by \cite{brown2020language} work with GPT-3, where strategically crafted prompts enabled the model to perform tasks with minimal prior examples. Beyond mere task instruction, recent studies have shifted towards understanding the semantic and contextual nuances in prompts, examining how subtle changes can lead to significantly different responses from the LLM. 

\textit{Ask-Me-Anything} \cite{arora2022ask} prompting introduced focusing on using multiple imperfect prompts and aggregating them to improve model performance, particularly in question-answering formats. Another one, \textit{Chain-of-Thought} method \cite{wei2023chainofthought}, where the model generates a series of intermediate reasoning steps to improve performance on complex tasks. Also, \textit{least-to-most prompting} \cite{zhou2023leasttomost} a novel strategy to break down complex problems into simpler subproblems, significantly enhancing the model's capability to tackle more challenging problems than those presented in the prompts. The effectiveness of explanation was explored \cite{lampinen-etal-2022-language}, finding that explanations can enhance LLM's learning capabilities on complex tasks. Furthermore, a catalog of prompt engineering techniques was examined with ChatGPT \cite{white2023prompt}, emphasizing the importance of prompt engineering in enhancing LLM applications in software development and education. It also highlighted that effective prompt design is crucial in improving LLM performance, particularly in coding practices and learning experiences. Lastly, \textit{Directional Stimulus Prompting} \cite{li2023guiding} presents a novel framework that uses a tunable policy model to generate auxiliary prompts, guiding LLMs towards specific desired outcomes. This diversity in prompting strategies underscores the rapidly evolving landscape of LLMs, offering multiple directions to harness their capabilities more effectively. 

\vspace{-0.07in}
\section{Principles} \label{main_principle}
\vspace{-0.03in}
\subsection{Motivation}

Since the quality of the responses generated by a pretrained and aligned LLM is directly relevant to the quality of the prompts or instructions provided by the users, it is essential to craft prompts that the LLM can comprehend and respond to effectively. The prompts delivered to an LLM serve as a way to program the interaction between a user and the LLM, enhancing its ability to address a diverse range of tasks. The primary focus of this work is on the methodology of crafting and customizing prompts to enhance output quality. This necessitates a comprehensive grasp of the functioning and behaviors of LLMs, their underlying mechanisms, and the principles governing their responses. In this work, we achieve this goal through elaborating 26 principles for comprehensive prompts in different scenarios and circumstances, examples are shown in Fig.~\ref{overview_examples}.

\vspace{-0.07in}
\subsection{Overview}
\vspace{-0.03in}

The overview of principles is presented in Table~\ref{tab:principles}. According to their unique nature, we group them into five categories as in Table~\ref{tab:categories}: (1) Prompt Structure
and Clarity, e.g., {\em integrate the intended audience in the prompt such as the audience is an expert in the field}; (2) Specificity and Information, e.g., {\em Add to your prompt the following phrase ``Ensure that your answer is unbiased and does not rely on stereotypes.''}; (3) User Interaction and Engagement, e.g., {\em Allow the model to elicit precise details and requirements from you by asking you questions until he has enough information to provide the needed output ``From now on, I would like you to ask me questions to...''.} (4) Content and Language Style, e.g., {\em No need to be polite with LLM so there is no need to add phrases like ``please'', ``if you don't mind'', ``thank you'', ``I would like to'', etc., and get straight to the point}; (5) Complex Tasks and Coding Prompts, e.g., {\em Break down complex tasks into a sequence of simpler prompts in an interactive conversation.}

\include{./tables/principles}

\include{./tables/principles_group}

\clearpage

\subsection{Design Principles}

In this study, a number of guiding principles are established for formulating prompts and instructions to elicit high-quality responses from pre-trained large language models: 

\noindent{\bf Conciseness and Clarity:} Generally, overly verbose or ambiguous prompts can confuse the model or lead to irrelevant responses. Thus, the prompt should be concise, avoiding unnecessary information that does not contribute to the task while being specific enough to guide the model. This is the basic principle guidance for prompt engineering.

\noindent{\bf Contextual Relevance:} The prompt must provide relevant context that helps the model understand the background and domain of the task. Including keywords, domain-specific terminology, or situational descriptions can anchor the model's responses in the correct context. We highlight this design philosophy in our presented principles.

\noindent{\bf Task Alignment:} The prompt should be closely aligned with the task at hand, using language and structure that clearly indicate the nature of the task to the model. This may involve phrasing the prompt as a question, a command, or a fill-in-the-blank statement that fits the task's expected input and output format.

\noindent{\bf Example Demonstrations:} For more complex tasks, including examples within the prompt can demonstrate the desired format or type of response. This often involves showing input-output pairs, especially in ``few-shot'' or ``zero-shot'' learning scenarios.

\noindent{\bf Avoiding Bias:} Prompts should be designed to minimize the activation of biases inherent in the model due to its training data. Use neutral language and be mindful of potential ethical implications, especially for sensitive topics.

\noindent{\bf Incremental Prompting:} For tasks that require a sequence of steps, prompts can be structured to guide the model through the process incrementally. Break down the task into a series of prompts that build upon each other, guiding the model step-by-step. Also, prompts should be adjustable based on the performance of the model and iterative feedback, i.e., it needs to be well prepared to refine the prompt based on initial outputs and model behaviors. Moreover, prompts should be adjustable based on the performance and response of the model, and iterative human feedback and preference.

Finally, more advanced prompts may incorporate programming-like logic to achieve complex tasks. For instance, use of conditional statements, logical operators, or even pseudo-code within the prompt to guide the model's reasoning process. The design of prompts is an evolving field, especially as LLMs become more sophisticated. As researchers continue to explore the limits of what can be achieved through prompt engineering, these principles will likely be refined and expanded.

\section{Experiments} \label{exp}

\subsection{Setup and Implementation Details}

All our evaluation is performed on ATLAS~\cite{ATLAS}, a manually crafted benchmark for principled prompt evaluation. 
It contains a standard subset featuring questions across various domains, along with a challenging subset dedicated to reasoning and other complex tasks. In our evaluation, we utilize a single response for each question. 
For each principle and the challenging subset, it contains 20 human-selected questions with and without the principled prompts. Similar to~\cite{alpaca_eval,zheng2023judging}, we compare each pair of responses from the same instructions with and without principles, and evaluate the various scales of LLM outputs by human evaluation. 

\begin{figure}[t]
  \centering
    \includegraphics[width=0.98\linewidth]{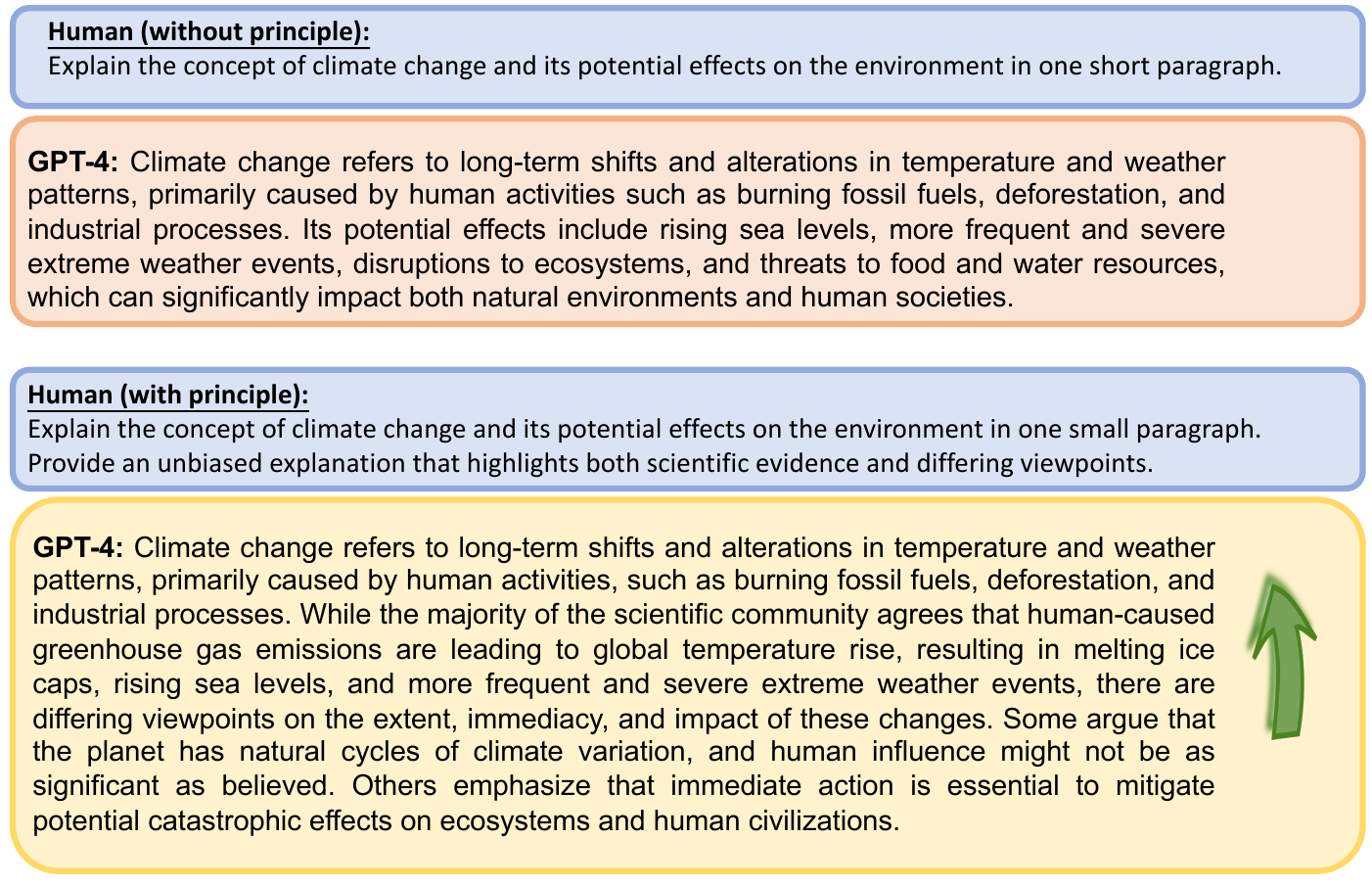}
  \vspace{-0.15in}
  \caption{Boosting example of LLM response after using the principle 13 on prompts.}
  \label{boosting_example}
  \vspace{-0.1in}
\end{figure}

\begin{figure}[t]
  \centering
    \includegraphics[width=0.94\linewidth]{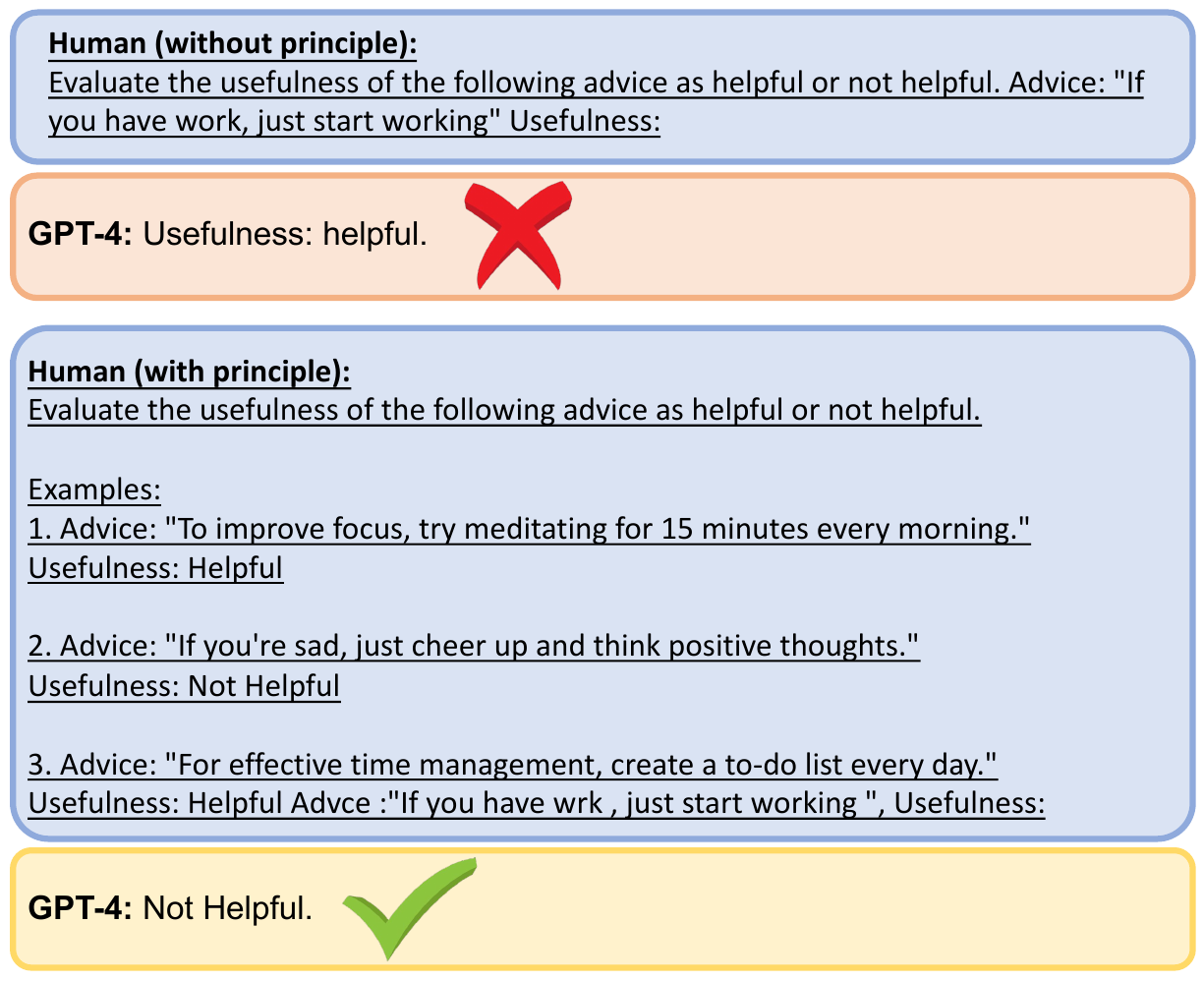}
  \vspace{-0.15in}
  \caption{Correctness improvement example of LLM response after using the introduced principle 7 on prompts.}
  \label{correct_example}
  \vspace{-0.1in}
\end{figure}

\begin{figure}[t]
  \centering
    \includegraphics[width=0.95\linewidth]{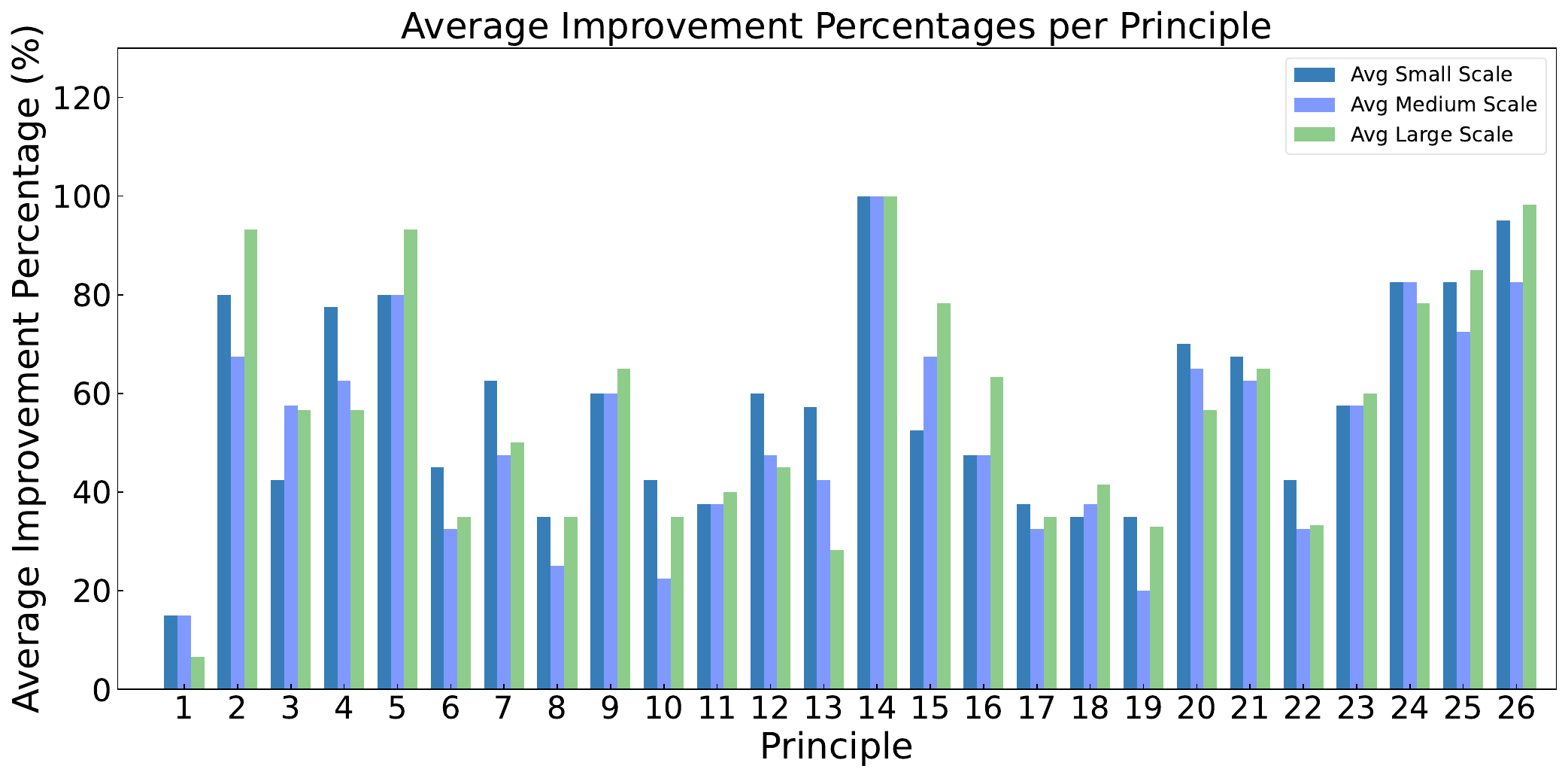}
  \vspace{-0.18in}
  \caption{Boosting of LLM response quality after employing the introduced principles on prompts. {\em small-scale} indicates the 7B models, {\em medium-scale} indicates the 13B models and {\em large-scale} indicates the 70B and GPT-3.5/4 models.}
  \label{improve_hist}
    \centering
    \includegraphics[width=0.95\linewidth]{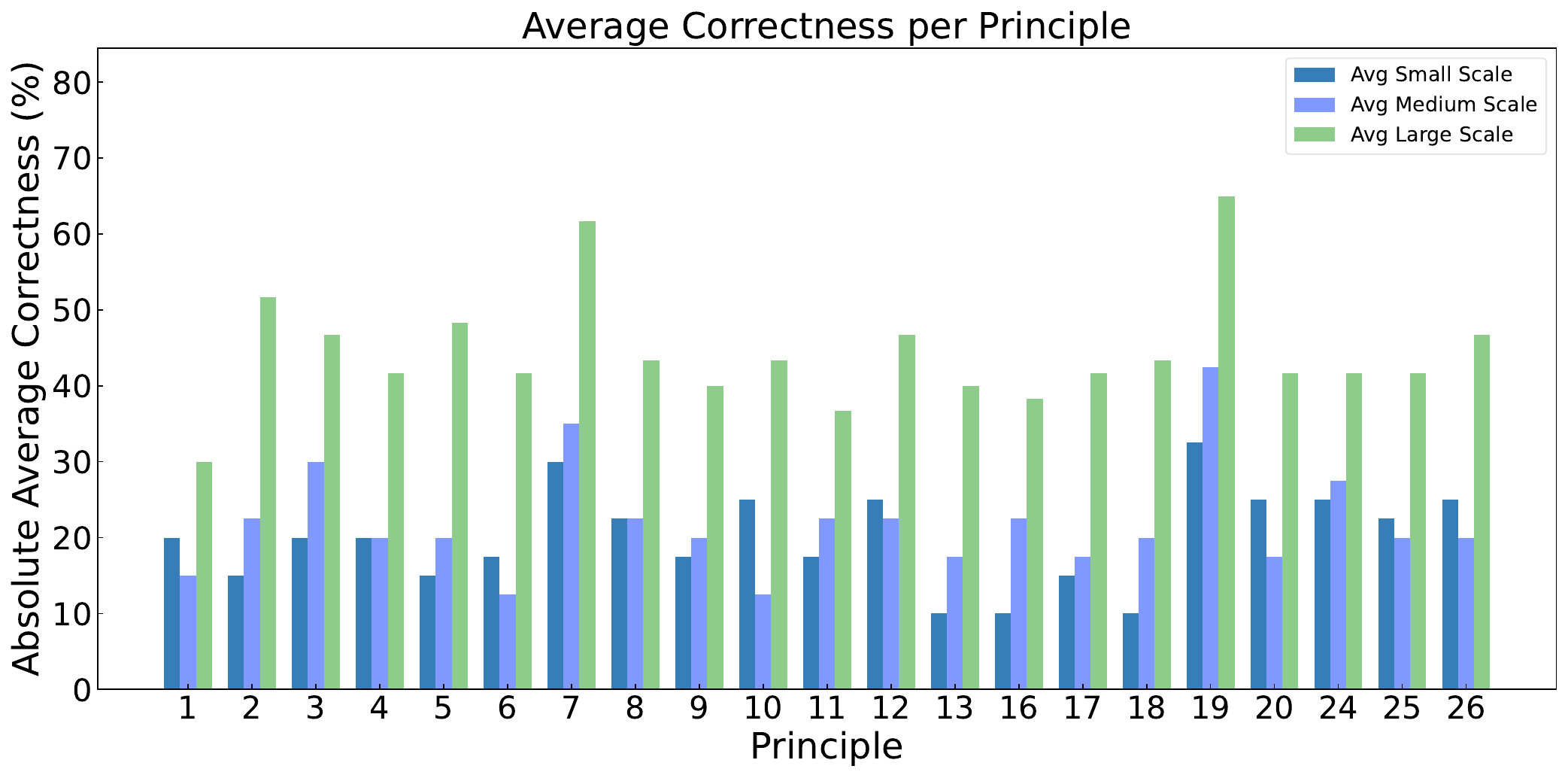}
  \vspace{-0.18in}
  \caption{Absolute correctness of LLM response quality after employing the introduced principles on prompts. {\em small-scale} indicates the 7B models, {\em medium-scale} indicates the 13B models and {\em large-scale} indicates the 70B and GPT-3.5/4 models.}
  \label{correct_hish}
  \vspace{-0.1in}
\end{figure}

\begin{figure}[t]
    \centering
    \includegraphics[width=0.95\linewidth]{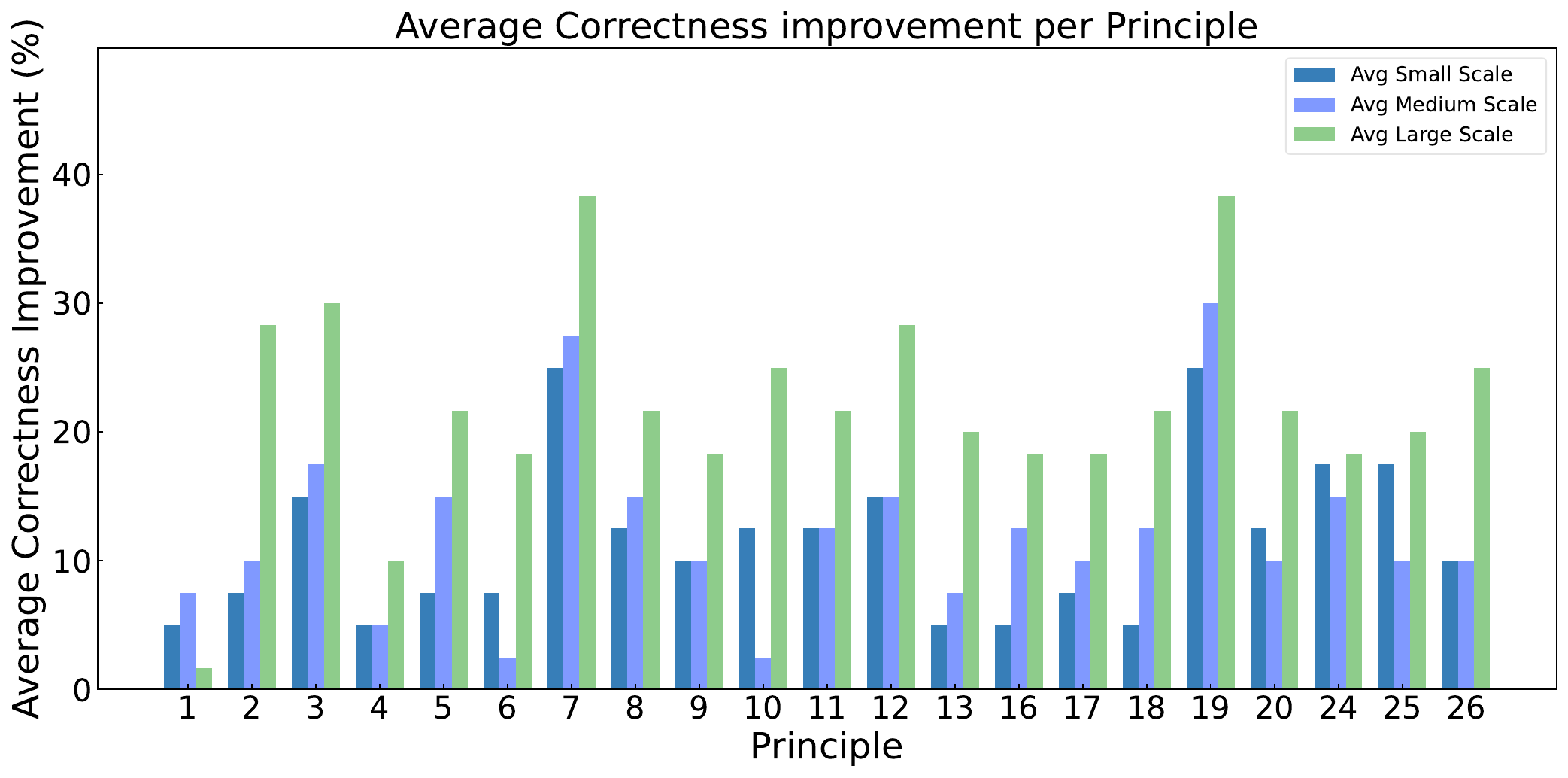}
  \vspace{-0.12in}
  \caption{Relative correctness improvement of LLM response quality after employing the introduced principles on prompts. {\em small-scale} indicates the 7B models, {\em medium-scale} indicates the 13B models and {\em large-scale} indicates the 70B and GPT-3.5/4 models.}
  \label{correct_hish_relative}
\end{figure}

\subsection{Models and Metrics}

We use instruction finetuned LLaMA-1-\{7, 13\}, LLaMA-2-\{7, 13\}, off-the-shelf LLaMA-2-70B-chat, GPT-3.5 (ChatGPT) and GPT-4 as our base models. We group these models into different scales: small-scale (7B models), medium-scale (13B) and large-scale (70B, GPT-3.5/4). We evaluate these models in two settings: {\bf Boosting} and {\bf Correctness}. They are employed together to provide a comprehensive understanding of a model’s performance. For correctness, we specifically utilize complex reasoning tasks to accurately gauge the precision of the models' outputs, contrasting with our evaluation for boosting, where simpler tasks are employed to effectively measure quality improvements. This distinction ensures a better reflection of the true capabilities for different scales of models and the effect of the principles for prompts. 
Since we use questions that typically involve complex reasoning tasks for correctness, some principles are not applicable including principles 14, 15, 21, 22, 23. For instance, ``{\em Suppose \( a \) and \( b \) are positive real numbers with \( a > b \) and \( ab = 8 \). Find the minimum value of \( \frac{a^2 + b^2}{a - b} \).}''

\begin{itemize}
     \item {\bf Boosting.} The result of {\em boosting} refers to the percentage increase in response quality across a set of questions when the proposed principles are applied. We assess the enhancement in the quality of responses from different LLMs via human evaluation after applying the outlined prompt principles. The original, unmodified prompts act as a baseline for measuring this enhancement. Demonstrating {\em boosting} confirms that a model's performance has improved due to the use of structured, principled instructions, as shown in Fig.~\ref{boosting_example}. 

     \item {\bf Correctness.} The concept of {\em correctness} refers to the precision of the model's outputs or responses, ensuring they are accurate, relevant, and devoid of errors. We consider both absolute and relative correctness accuracy. Human evaluators are utilized to gauge this aspect, which is crucial for verifying the model's accuracy. Correctness is a testament to the model's ability to generate outputs that align with the expected standards of accuracy, as shown in Fig.~\ref{correct_example}.
\end{itemize}

\subsection{Results}

\begin{figure}[t]
  \centering
    \includegraphics[width=0.999\linewidth]{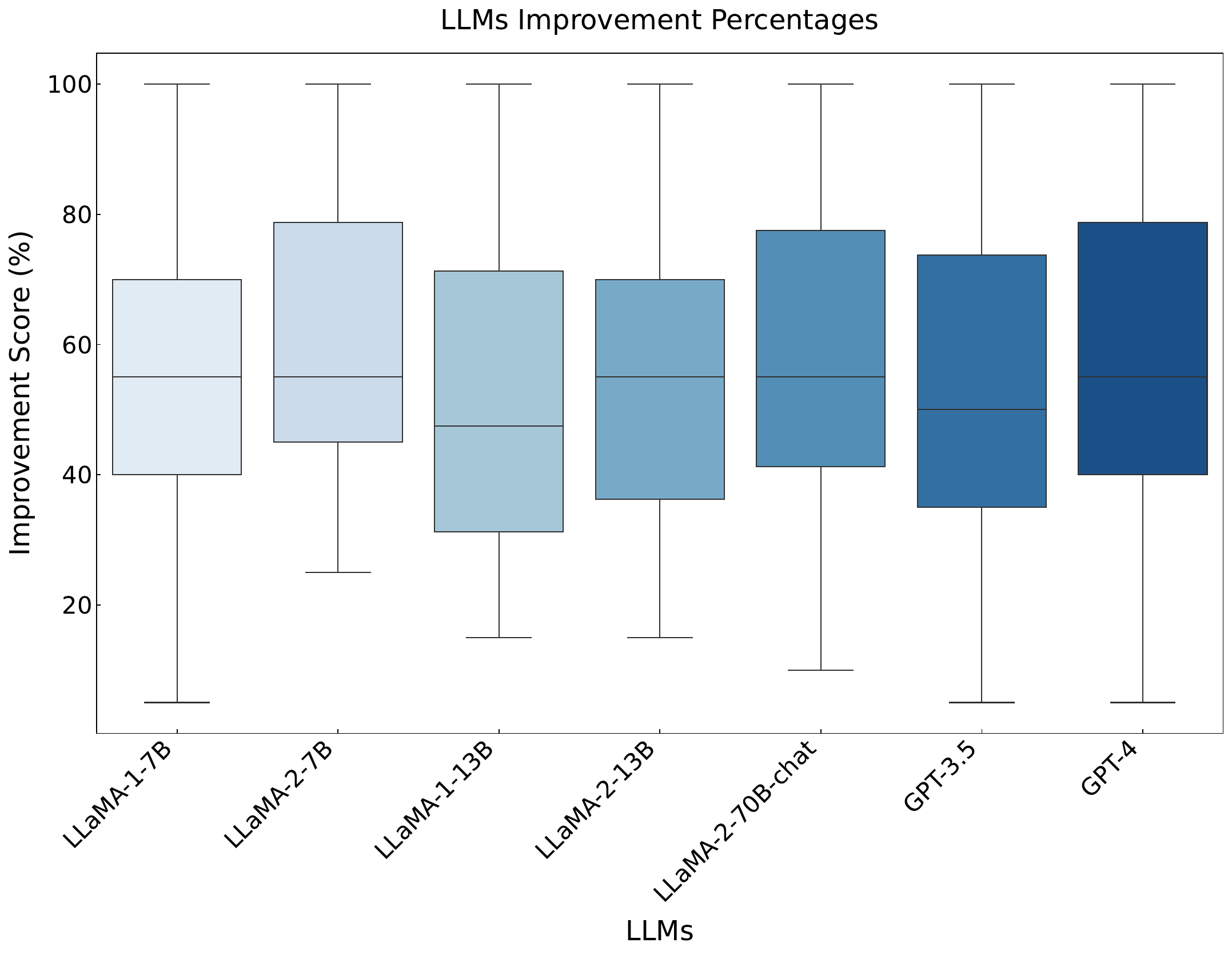}
  \vspace{-0.3in}
  \caption{Boosting score across various LLMs on the ATLAS dataset.}
  \label{individual_improvement}
  \vspace{0.05in}

  \centering
    \includegraphics[width=0.999\linewidth]{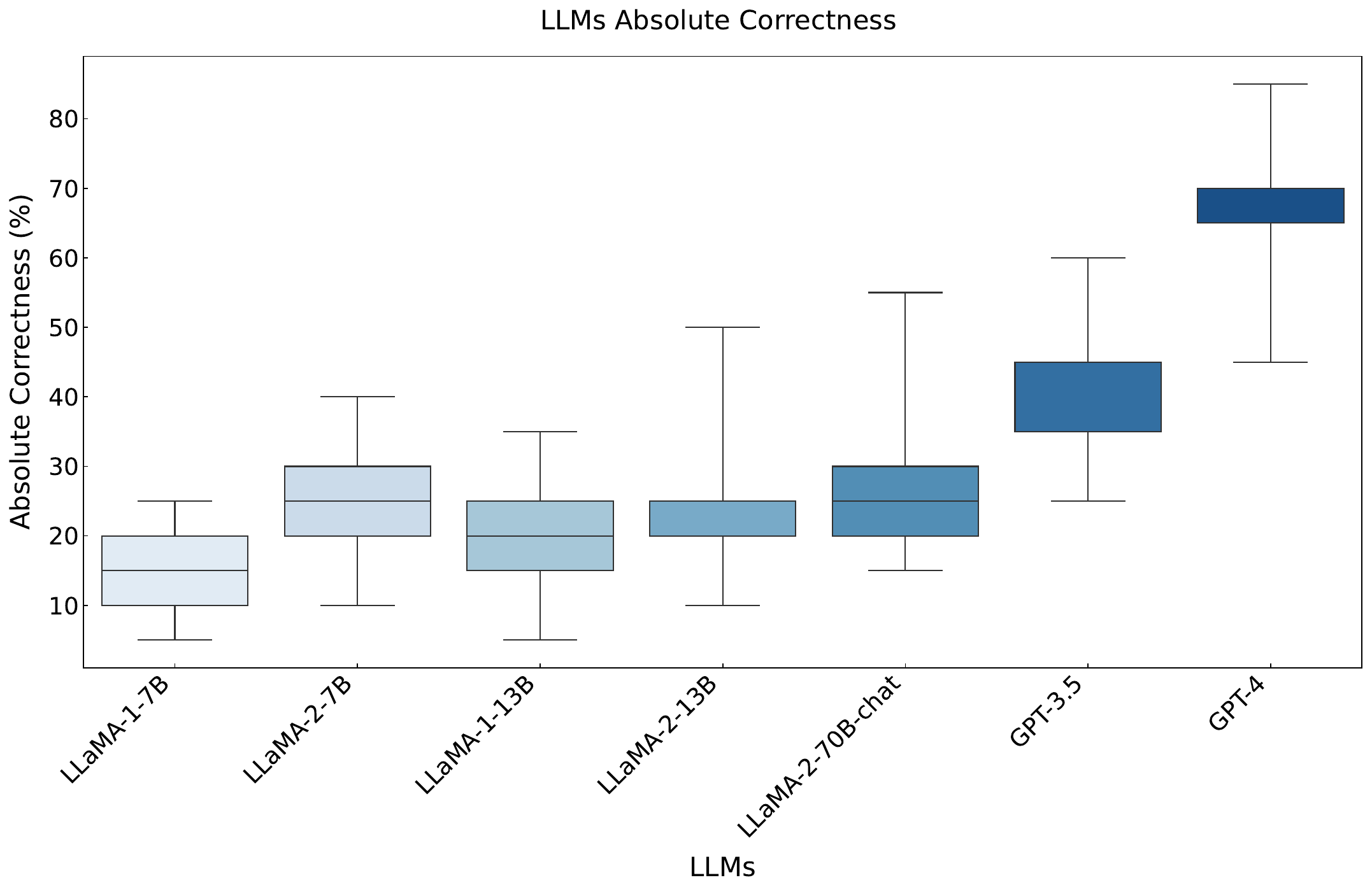}
  \vspace{-0.3in}
  \caption{Absolute correctness score on the ATLAS dataset.}
  \label{individual_correct}
\end{figure}

\begin{figure}[t]
  \centering
    \includegraphics[width=0.999\linewidth]{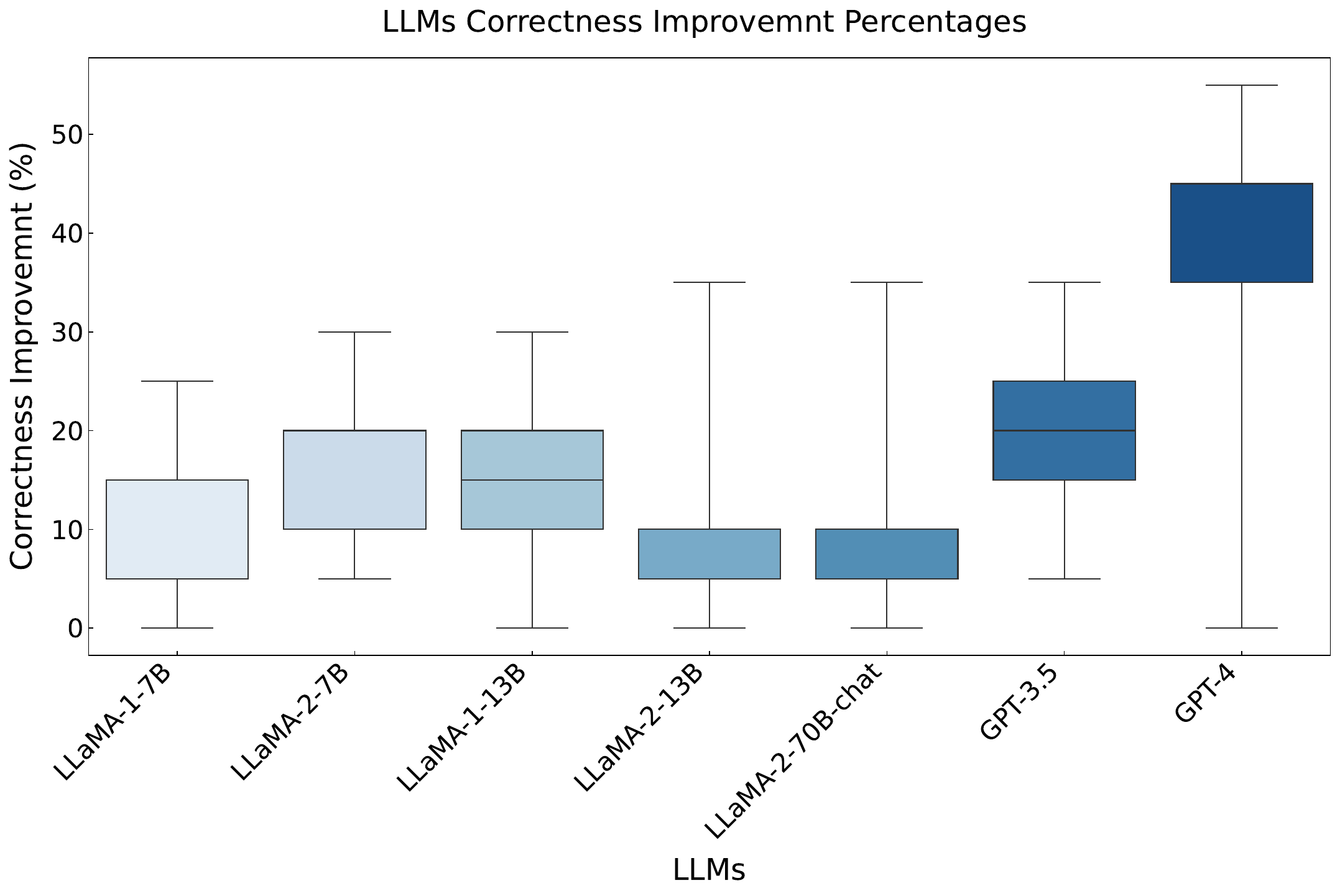}
  \vspace{-0.3in}
  \caption{Relative correctness improvement score on the ATLAS dataset.}
  \label{individual_correct_relative}
\end{figure}

\begin{figure}[t]
  \centering
    \includegraphics[width=0.9\linewidth]{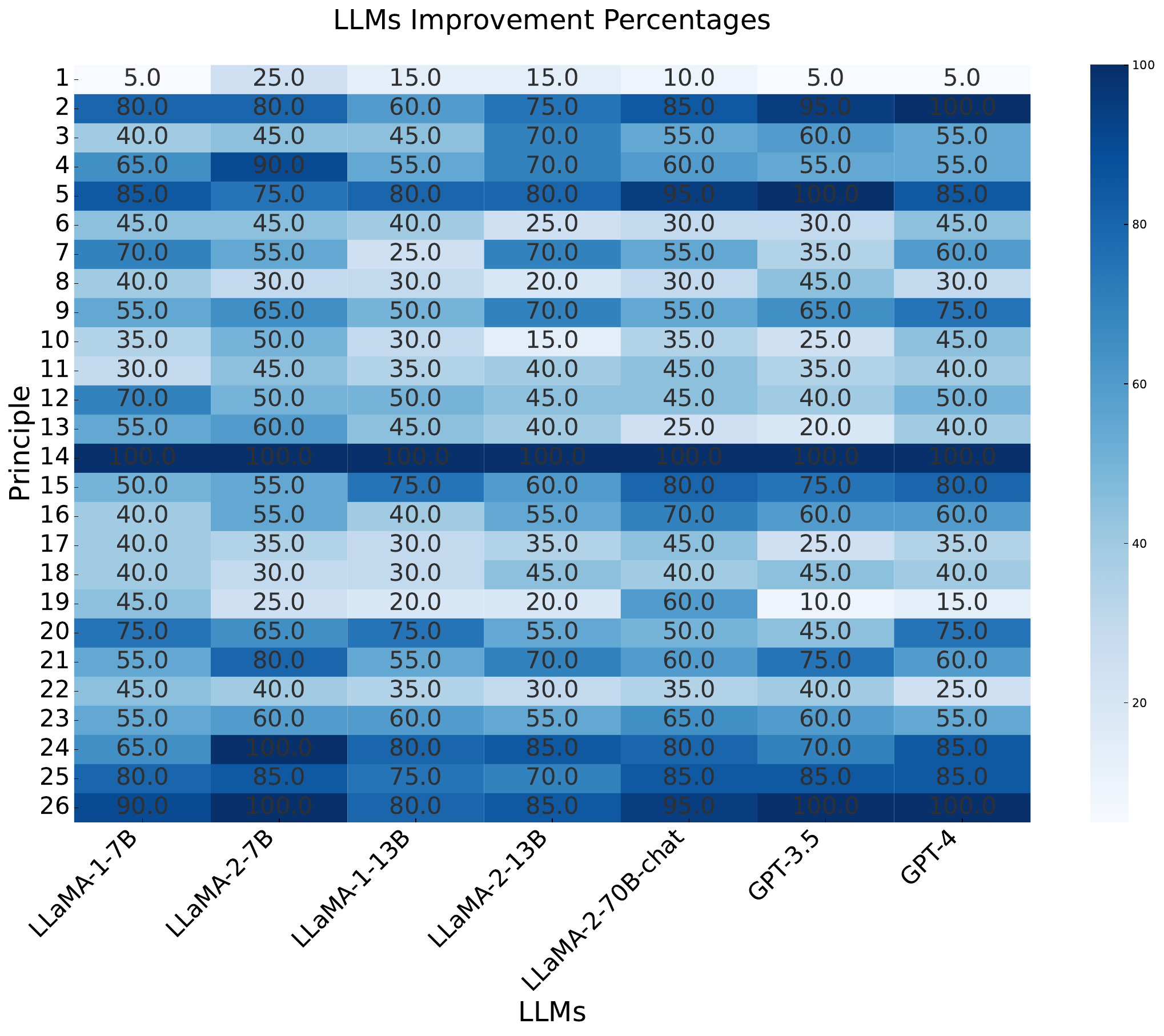}
  \vspace{-0.1in}
  \caption{Illustration of heatmap for LLMs boosting percentages.}
  \label{heatmap_boost}
  \vspace{-0.1in}
\end{figure}

\begin{figure}[t]
  \centering
    \includegraphics[width=0.9\linewidth]{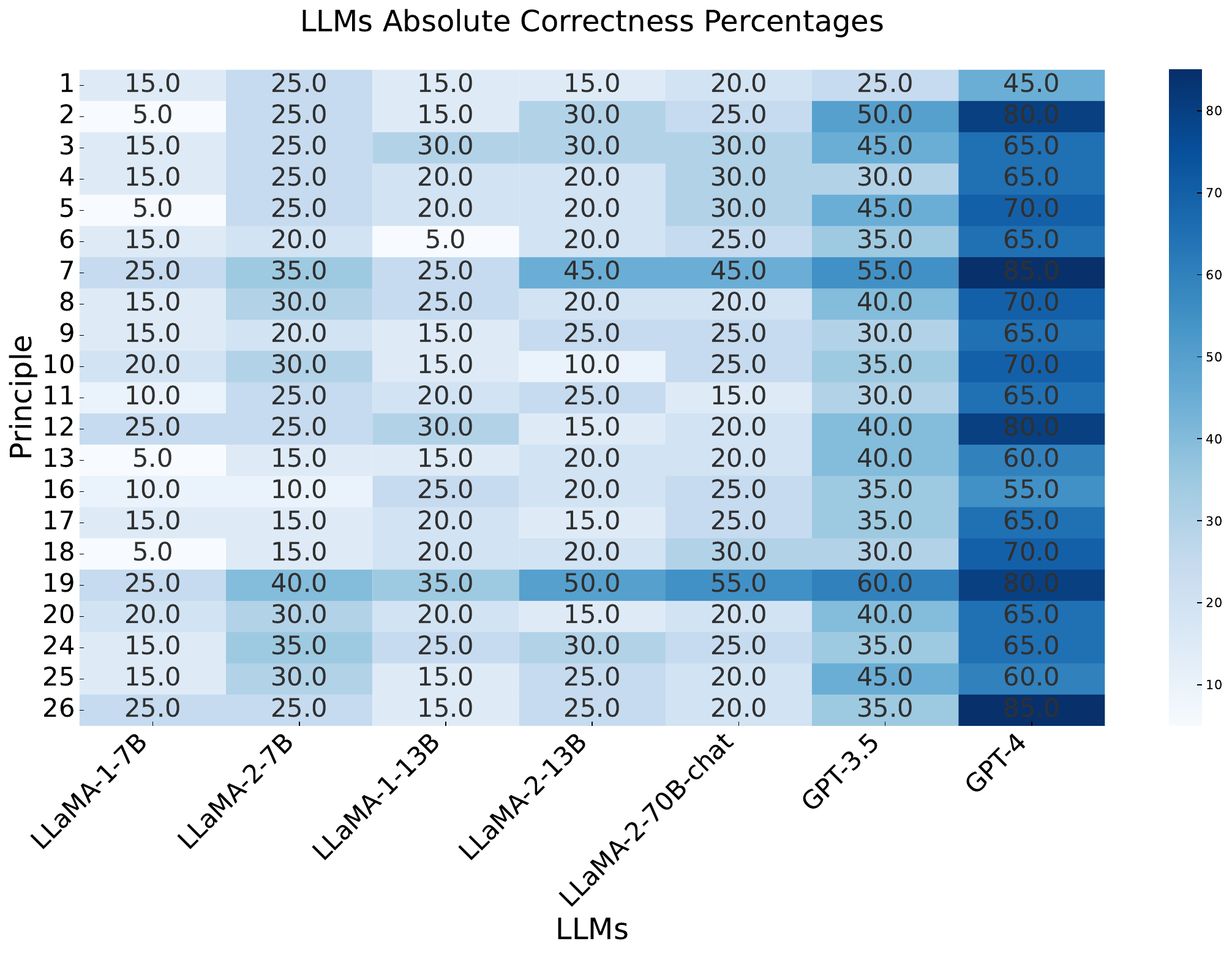}
  \vspace{-0.1in}
  \caption{Illustration of heatmap for absolute correctness percentages.}
  \label{heatmap_correct}
  \vspace{-0.1in}
\end{figure}

\begin{figure}[t]
  \centering
    \includegraphics[width=0.9\linewidth]{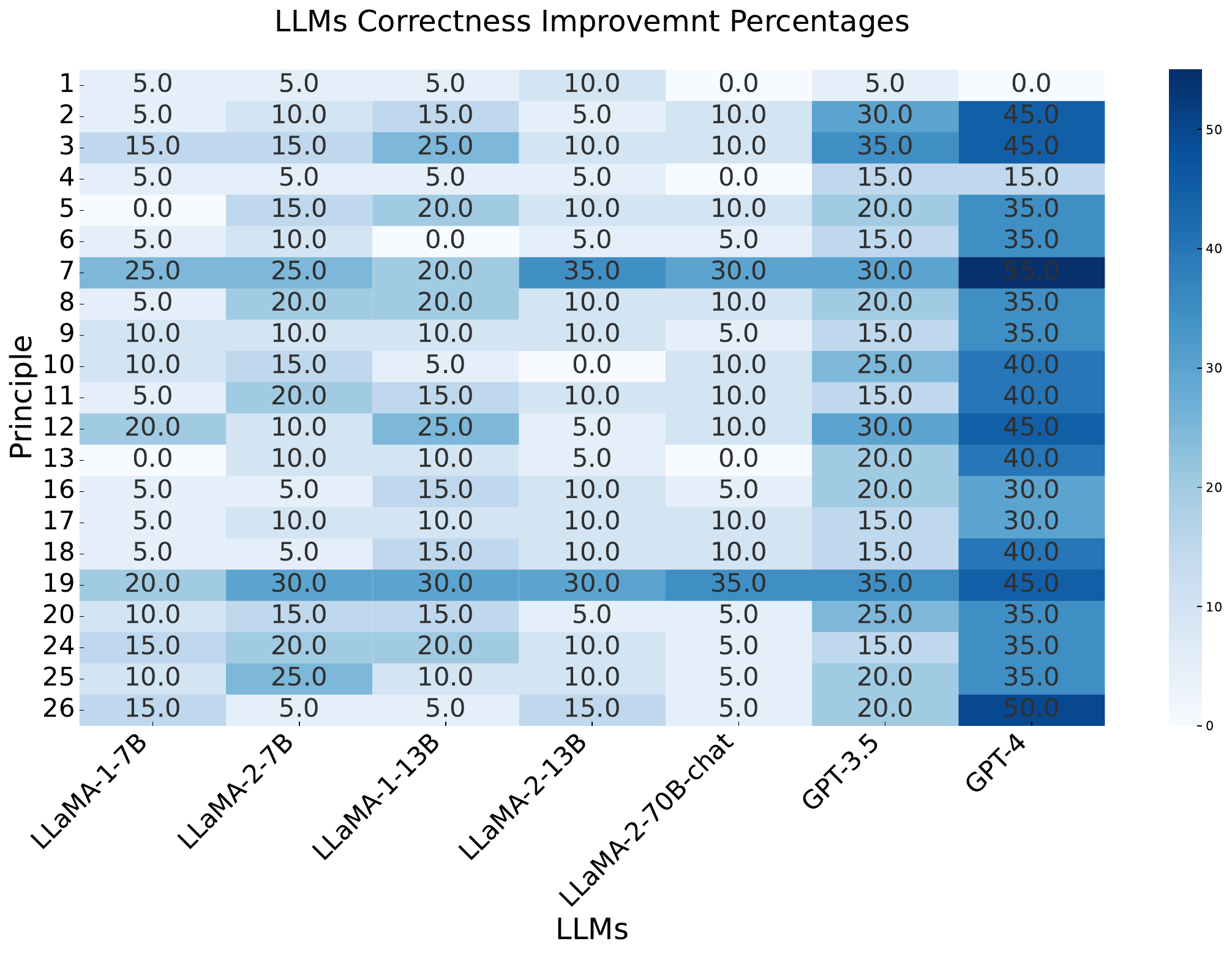}
  \vspace{-0.1in}
  \caption{Illustration of heatmap for relative correctness improvement percentages.}
  \label{heatmap_correct_relative}
  \vspace{-0.1in}
\end{figure}

\subsubsection{Results on small, medium and large-scale LLMs}

\noindent{\bf Boosting.} The results of improvement after employing the introduced principles are shown in Fig.~\ref{improve_hist}. Generally, all principles can bring a significant improvement on the three scales of LLMs. In the cases of principles 2, 5, 15, 16, 25 and 26, the large-scale models get the most improvement by the principled prompts. Particularly, for principle 14, as shown in Fig.~\ref{improve_hist}, it has improved all questions it is applied to.

\noindent{\bf Correctness.} (1) Absolute accuracy: we examine the absolute performance when employing the principles on various scales of models. Generally, these models achieve 20\%$\sim$40\% accuracy on the averaged performance, as shown in Fig.~\ref{correct_hish}. In particular, for small and medium scale models, the accuracy can basically reach between 10\% and 40\%, and for large models, the accuracy can reach more than 40\%. (2) Relative accuracy: Fig.~\ref{correct_hish_relative} illustrates that applying the principles generally leads to a performance increase of over 10\% across different models on average. For larger models, this enhancement can surpass 20\%.

\subsubsection{Results on individual LLMs}

\noindent{\bf Boosting.}  Fig.~\ref{individual_improvement} illustrates the improvement of response quality on individual model and principle after using the revised prompts. On average, there is a stable 50\% improvement across different LLMs. Fig.~\ref{heatmap_boost} further provides the detailed results of improvement for each principle with different LLMs.

\noindent{\bf Correctness.} Fig.~\ref{individual_correct} illustrates the absolute correctness accuracy and Fig.~\ref{individual_correct_relative} shows the relative enhancements in accuracy across different sizes of LLMs. From LLaMA-2-13B, LLaMA-2-70B-chat to GPT-3.5 and GPT-4, there is a noticeable trend: the larger the model, the greater the increase in correctness improvement. Fig.~\ref{heatmap_correct} and Fig.~\ref{heatmap_correct_relative} further present the absolute and relative correctness enhancements by each principle.

\subsubsection{More examples on various scales of LLMs}

\begin{figure}[t]
  \centering
    \includegraphics[width=0.98\linewidth]{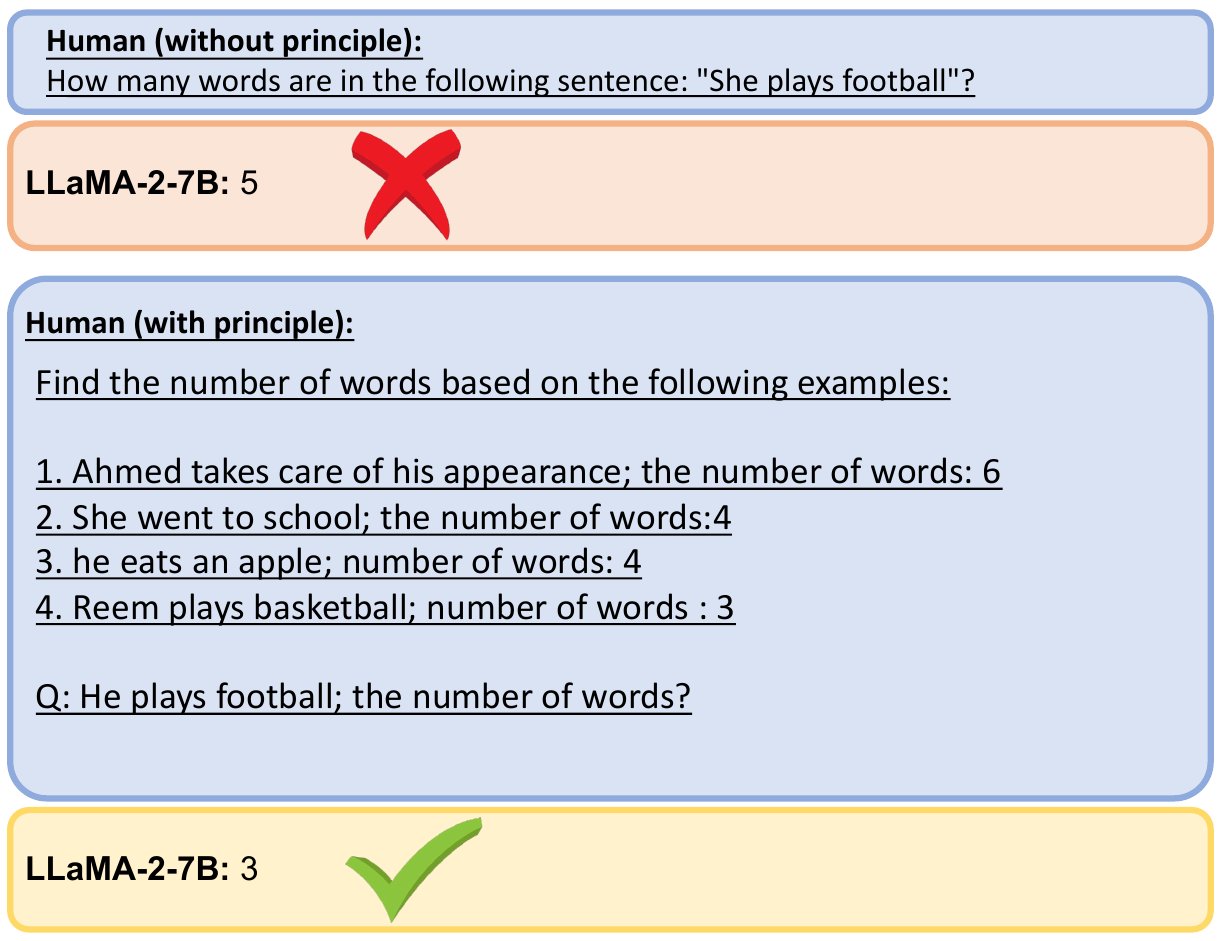}
  \vspace{-0.15in}
  \caption{Correctness improvement on small-scale LLaMA-2-7B model after using the introduced principle on prompts.}
  \label{small_example1}
  \vspace{-0.1in}
\end{figure}

\begin{figure}[t]
  \centering
    \includegraphics[width=0.98\linewidth]{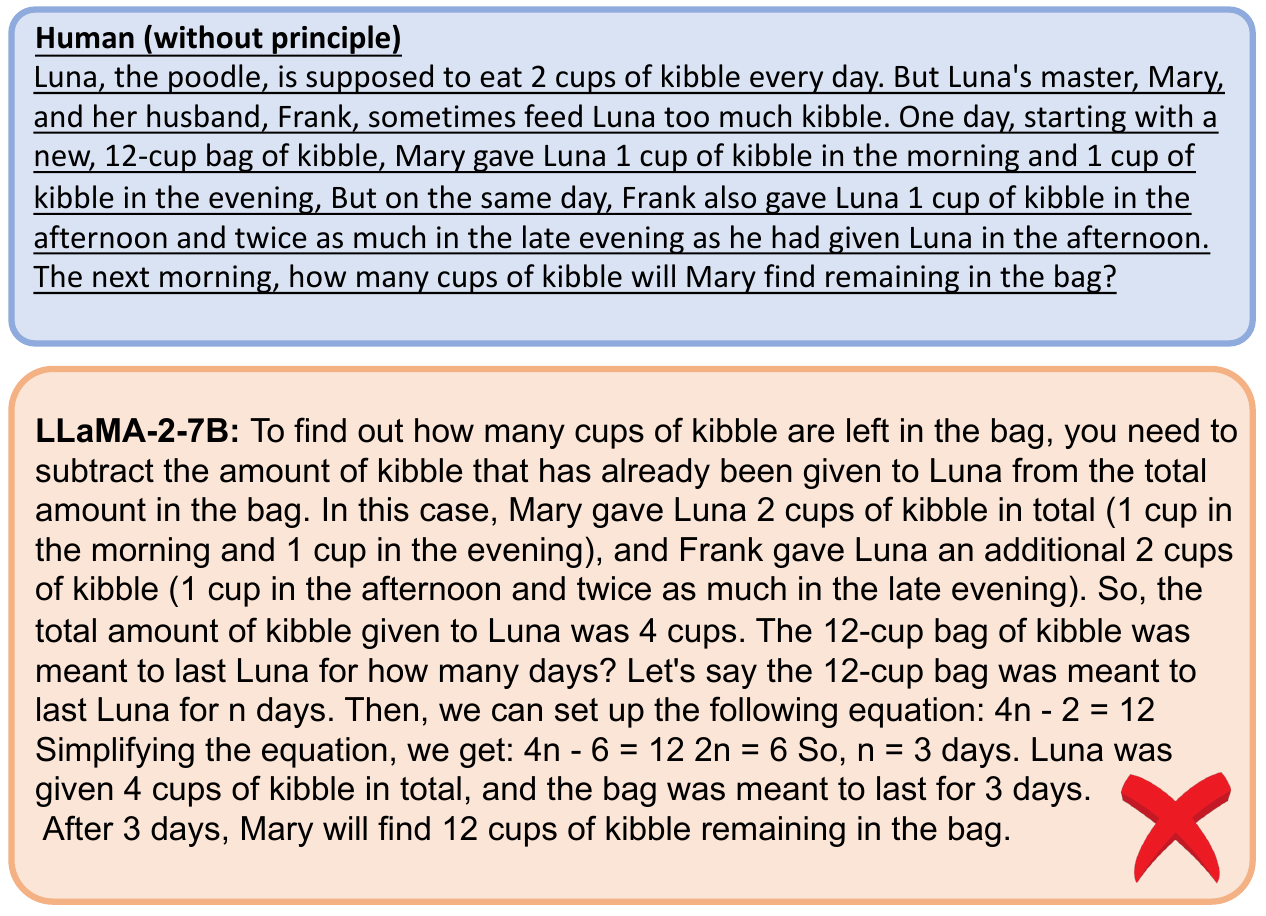}
    \includegraphics[width=0.98\linewidth]{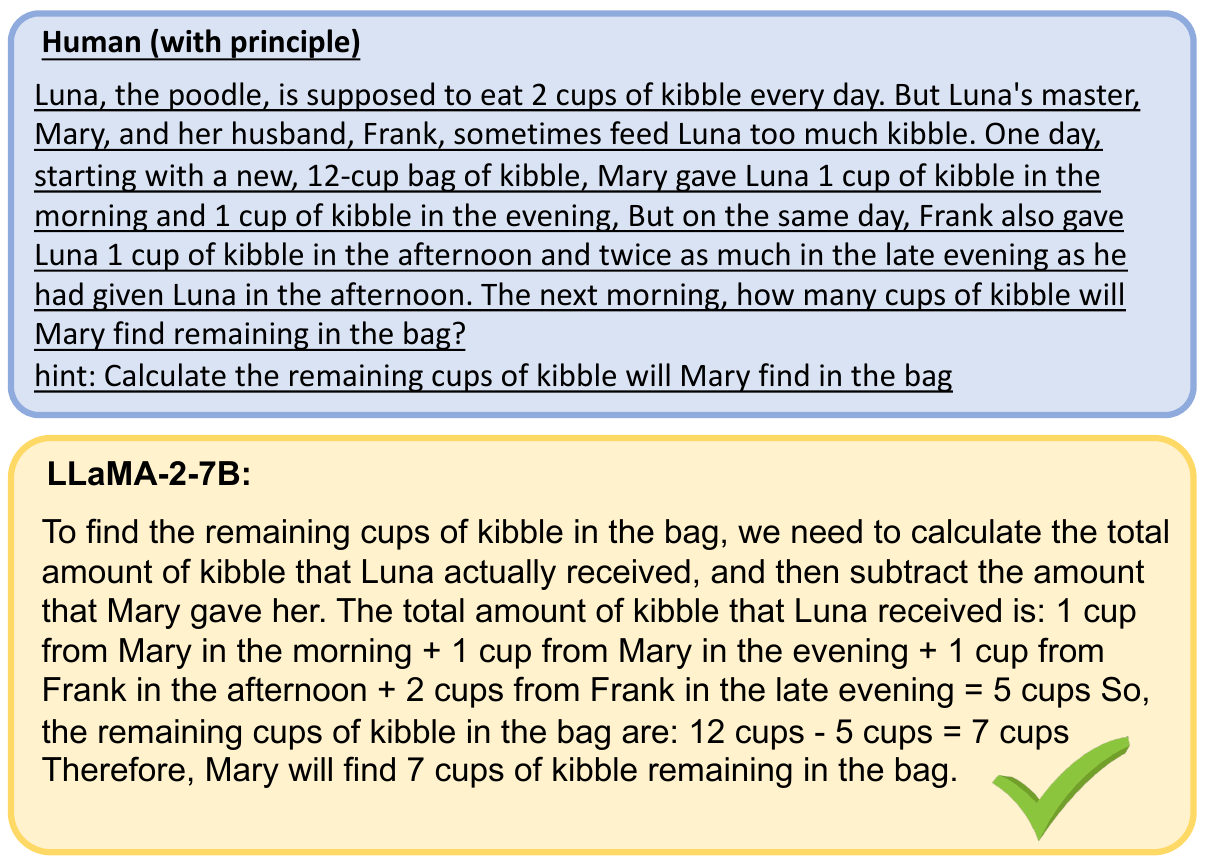}
  \vspace{-0.15in}
  \caption{Correctness improvement on small-scale LLaMA-2-7B model after using the introduced principle on prompts.}
  \label{small_example2}
  \vspace{-0.1in}
\end{figure}

\begin{figure}[t]
  \centering
    \includegraphics[width=0.98\linewidth]{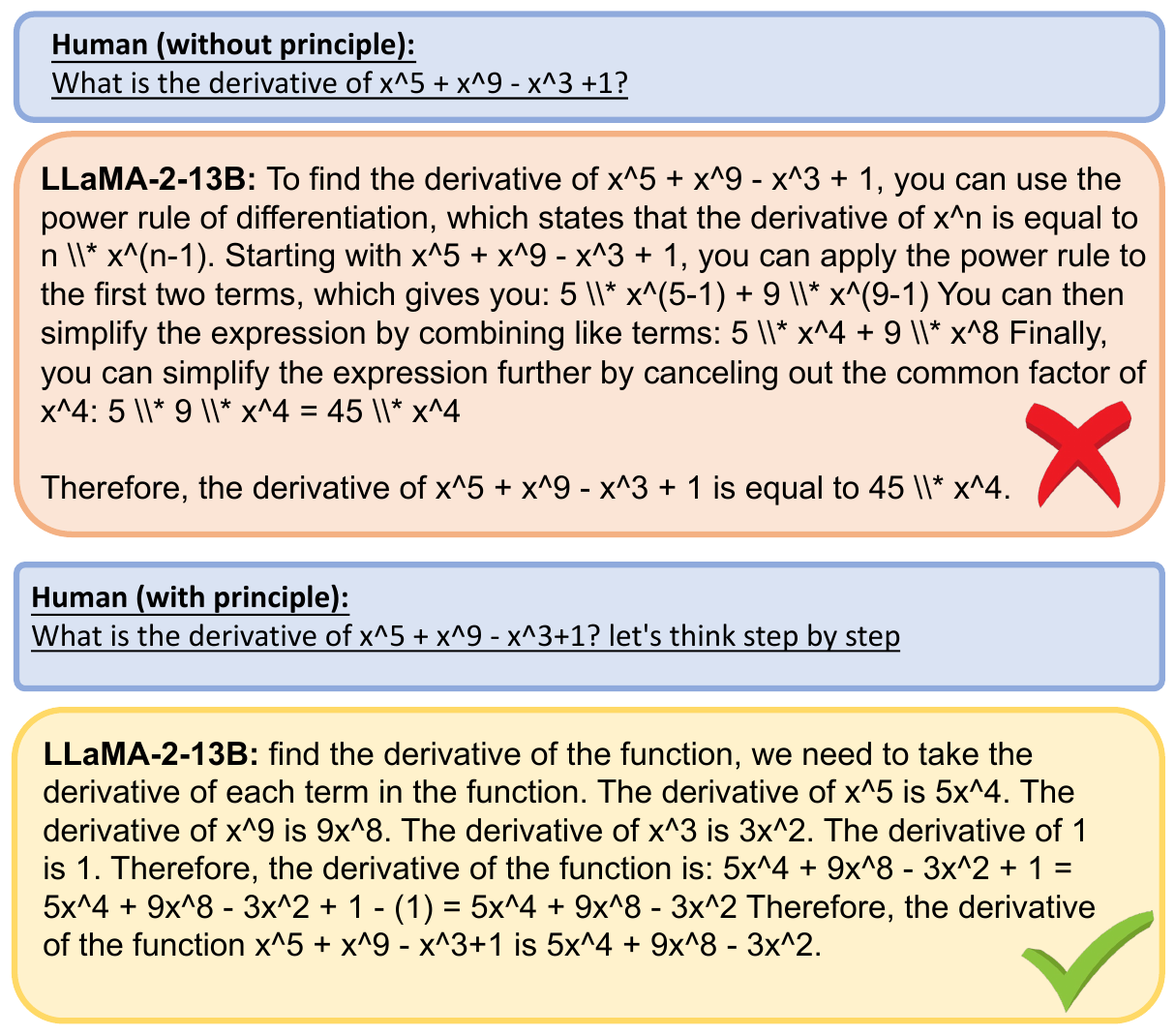}
  \vspace{-0.15in}
  \caption{Correctness improvement on medium-scale LLaMA-2-13B model after using the introduced principle on prompts.}
  \label{medium_example1}
  \vspace{-0.1in}
\end{figure}

\begin{figure}[t]
  \centering
    \includegraphics[width=0.98\linewidth]{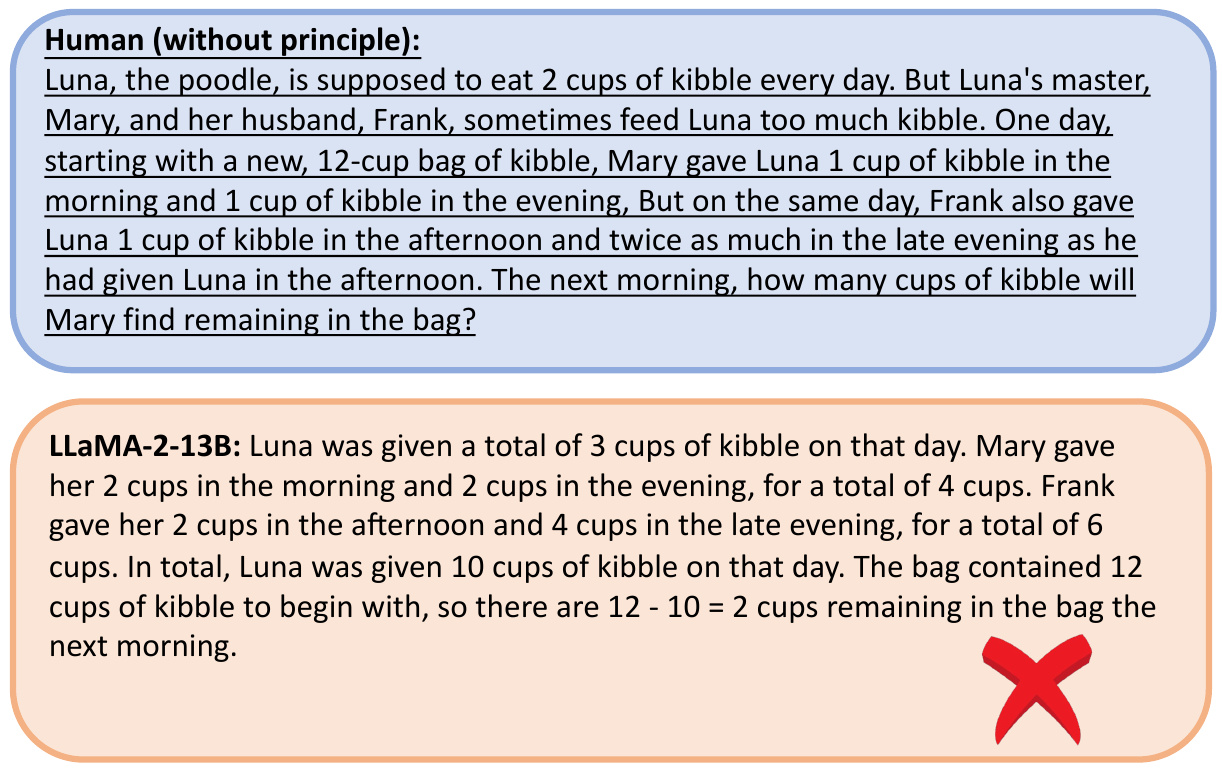}
    \includegraphics[width=0.98\linewidth]{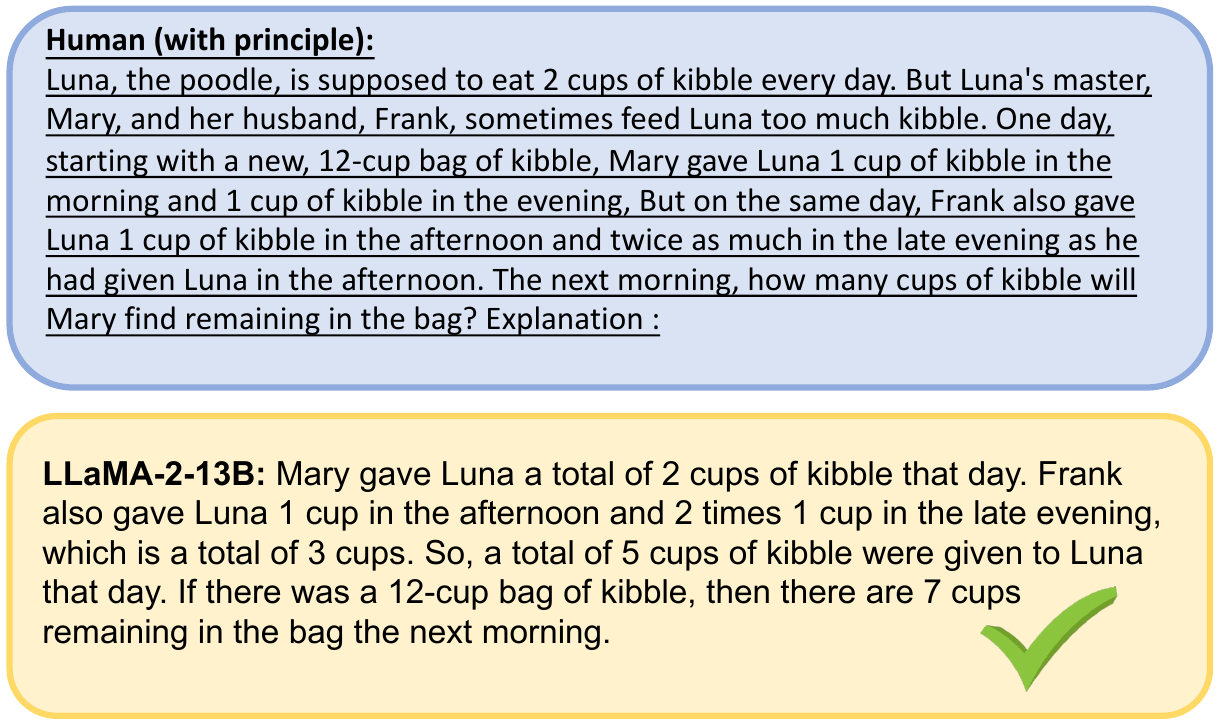}
  \vspace{-0.15in}
  \caption{Correctness improvement on medium-scale LLaMA-2-13B model after using the introduced principle on prompts.}
  \label{medium_example2}
  \vspace{-0.1in}
\end{figure}

We present additional examples for both small and medium-scale LLMs, as illustrated in Fig.~\ref{small_example1} and \ref{small_example2} for the small-scale LLaMA-2-7B, and Fig.~\ref{medium_example1} and \ref{medium_example2} for the medium-scale LLaMA-2-13B. Empirically, the use of the proposed principles on prompts has demonstrably enhanced the accuracy of the responses generated by these models.

\vspace{-0.08in}
\section{Conclusion}
\vspace{-0.05in}

We presented 26 principles through an exhaustive analysis that enhances the LLM ability to focus on the crucial elements of the input context, leading to the generation of quality responses. By guiding the LLM with these meticulously crafted principles before the input is processed, we can encourage the model towards producing better responses. Our empirical results demonstrate that this strategy can effectively reformulate contexts that might otherwise compromise the quality of the output, thereby enhancing the relevance, brevity, and objectivity of the responses.

\noindent{There are numerous directions for future exploration. In our experiments, we utilized a constrained shot prompting approach to apply these principles. There is potential to refine our base models to align with our principled instructions further with alternative strategies, such as fine-tuning, reinforcement learning, direct preference optimization, or different prompting methods using our generated dataset. Moreover, the strategies that prove successful could be integrated into standard LLM operations, for instance, by fine-tuning with the original/principled prompts as inputs and the polished, principled responses as targets for training.}

\clearpage

\section{Limitations and Discussion}

While the proposed 26 principles are designed to improve and enhance the quality of responses of LLMs across a diverse array of queries, the effectiveness of these principles may diminish when dealing with questions that are very complex or highly specialized. This limitation can mainly depend on the reasoning capabilities and training of each model. To address these variations, we have tested the principles across different scales to measure their effectiveness comprehensively.

Despite our efforts in evaluating these principles on seven distinct language models, it is crucial to acknowledge that models with architectures different from those tested might respond in different ways to these principles. Additionally, our assessment of improvement and correctness percentages was based on a limited selection of questions. Expanding the question set in future research could yield more generalized findings and offer deeper insights into the applicability of each principle. Furthermore, the criteria and results may vary across various personnel assessments on the model responses.

\clearpage

{\small
\bibliographystyle{ieee_fullname}
\bibliography{my}
}

\newpage

\end{document}

%% file: tables/principles.tex
\begin{table*}[h]
\vspace{-1.2in}
\hspace{-0.4in}
\fontsize{7.8pt}{11.5}\selectfont
\setlength{\tabcolsep}{5pt} 
\begin{tabular}{|c|p{12.5cm}|l|}
\hline
\textbf{\#Principle} & \hspace{1.5in} \textbf{Prompt Principle for Instructions}             \\ \hline
1 & \begin{tabular}[c]{@{}l@{}} If you prefer more concise answers, no need to be polite with LLM so there is no need to add phrases like \\``please'', ``if you don't mind'', ``thank you'', ``I would like to'', etc., and get straight to the point. \end{tabular}  \\ \hline
2           & Integrate the intended audience in the prompt, e.g., the audience is an expert in the field.     
\\ \hline
3      & Break down complex tasks into a sequence of simpler prompts in an interactive conversation.                    \\ \hline
4        & Employ affirmative directives such as `{\em do,}' while steering clear of negative language like `{\em don’t}'.  
\\ \hline
5          & \begin{tabular}[c]{@{}l@{}}
When you need clarity or a deeper understanding of a topic, idea, or any piece of information, utilize the \\following prompts:\\ o Explain [insert specific topic] in simple terms. \\ o Explain to me like I'm 11 years old. \\ o Explain to me as if  I'm a beginner in [field]. \\ o Write the [essay/text/paragraph]  using simple  English like you're explaining something to a 5-year-old. \end{tabular}                                                                                                         \\ \hline
6 & Add ``I'm going to tip \$xxx for a better solution!''  \\ \hline
7           &  Implement example-driven prompting (Use few-shot prompting).       
\\ \hline
8       & \begin{tabular}[c]{@{}l@{}}
When formatting your prompt, start with `\#\#\#Instruction\#\#\#', followed by either `\#\#\#Example\#\#\#' \\or `\#\#\#Question\#\#\#' if relevant. Subsequently, present your content. Use one or more \\ line breaks to separate instructions, examples, questions, context, and input data. \end{tabular} 
\\ \hline
9           & Incorporate   the following phrases: ``Your task is'' and ``You MUST''.                                                    \\ \hline
10          & Incorporate   the following phrases: ``You will be penalized''.                                                             \\ \hline
11          & Use the phrase "Answer a question given in a natural, human-like manner" in your prompts.                                                                                                                    \\ \hline
12        & Use leading words like writing ``think step by step''.                                                                                            \\ \hline
13       & \begin{tabular}[c]{@{}l@{}}Add to your prompt the following phrase ``Ensure that your answer is unbiased and avoids relying on stereotypes.''

\end{tabular}                                          \\ \hline
14         & \begin{tabular}[c]{@{}l@{}}Allow the model to elicit precise details and requirements from you by asking you questions until he has \\ enough information to provide the needed output (for example, ``From now on, I would like you to ask me \\ questions to ...'').\end{tabular}     
\\ \hline
15     & \begin{tabular}[c]{@{}l@{}}To inquire about a specific topic or idea or any information and you want to test your understanding, you can use \\the following phrase: ``Teach me any [theorem/topic/rule name] and include a test at the end, and let me know if \\ my answers are correct after I respond, without providing the answers beforehand.''

\end{tabular}                                                                                             \\ \hline
16    & Assign a role to the large language models.                                                                                                                                                                                                                  \\ \hline
17       & Use Delimiters.                                                                                                                                                                                                               \\ \hline
18    & Repeat a specific word or phrase multiple times within a prompt.                                                            
\\ \hline
19 & Combine Chain-of-thought (CoT) with few-Shot prompts.  \\ \hline
20       & \begin{tabular}[c]{@{}l@{}} Use output primers, which involve concluding your prompt with the beginning of the desired output.   Utilize output \\primers by ending your prompt with the start of the anticipated response.    \end{tabular}                                                                                                                                                                               \\ \hline
21  & \begin{tabular}[c]{@{}l@{}} To write an essay /text /paragraph /article or any type of text that should be detailed: ``Write a detailed [essay/text\\/paragraph] for me on [topic] in detail by adding all the information necessary''.\end{tabular} \\ \hline
22 & \begin{tabular}[c]{@{}l@{}} To correct/change specific text without changing its style: ``Try to revise every paragraph sent by users. You should \\only improve the user's grammar and vocabulary and make sure it sounds natural. You should maintain the original \\ writing style, ensuring that a formal paragraph remains formal.'' \end{tabular} \\ \hline

23  & \begin{tabular}[c]{@{}l@{}}When you have a complex coding prompt that may be in different files: 
``From now and on whenever you generate \\code that spans more than one file, generate a [programming language ] script that can be run to automatically \\create the specified files or make changes to existing files to insert the generated code. [your question]''. \end{tabular}\\ \hline
24   & \begin{tabular}{@{}p{12cm}@{}} When you want to initiate or continue a text using specific words, phrases, or sentences, utilize the following prompt: \newline o I'm providing you with the beginning [song lyrics/story/paragraph/essay...]: [Insert lyrics/words/sentence]. Finish it based on the words provided. Keep the flow consistent.
\end{tabular}
                        \\ \hline
25   & \begin{tabular}[c]{@{}l@{}}Clearly state the requirements that the model must follow in   order to produce content, \\ in the form of  the keywords, regulations, hint, or instructions\end{tabular}                                                                                                                                                                                            \\ \hline
26        & \begin{tabular}[c]{@{}l@{}}To write any text, such as an essay or paragraph, that is intended to be similar to a provided sample, include the \\following instructions: \\o Use the same language based on the provided paragraph[/title/text /essay/answer].\end{tabular}                                                                                                                   \\ \hline
\end{tabular}
\caption{Overview of 26 randomly ordered prompt principles.}
\label{tab:principles} 
\end{table*}

%% file: tables/principles_group.tex
\begin{table*}[h]
\vspace{-1.3in}
\fontsize{6.8pt}{8.5}\selectfont
\setlength{\tabcolsep}{5pt} 

\hspace{-0.5in}
\begin{tabular}{|c|p{10.77cm}|c|}
\hline
\bf Category &
 \hspace{2in} \bf Principles &
  \textbf{\#Principle} \\ \hline
\multirow{13}{*}{\begin{tabular}[c]{@{}c@{}}Prompt Structure \\ and Clarity\end{tabular}} &

  Integrate the intended audience in the prompt. &
  2 \\  

 &
 \begin{tabular}[c]{@{}l@{}}
  \\ Employ affirmative directives such as `do' while steering clear of negative language like `don't'. \end{tabular}& 
  4 \\ 
 &
 \begin{tabular}[c]{@{}l@{}}
  \\
  Use Leading words like writing ``think step by step.''   \end{tabular}&
  12 \\ 
 &
  \begin{tabular}{@{}p{10.7cm}@{}}
  \\ Use output primers, which involve concluding your prompt with the beginning of the desired output. \\ by ending your prompt with the start of the anticipated response.           \end{tabular}&
  20 \\ 
 &
 \begin{tabular}[c]{@{}l@{}}
  \\
  Use Delimiters.  \end{tabular}&
  17 \\
 &
   \begin{tabular}{@{}p{10.7cm}@{}} \\ When formatting your prompt, start with `\#\#\#Instruction\#\#\#', followed  by either `\#\#\#Example\#\#\#' or `\#\#\#Question\#\#\#' if relevant. Subsequently, present your content. Use one or more line breaks to separate instructions, examples, questions, context, and input data.\end{tabular} &
  8 \\ \hline

\multirow{26}{*}{\begin{tabular}[c]{@{}c@{}}Specificity and \\ Information\end{tabular}} &

\begin{tabular}[c]{@{}l@{}}
  Implement example-driven prompting (Use few-shot prompting). \end{tabular} &
  7 \\
 &
 \begin{tabular}{@{}p{10.7cm}@{}}
\\ When you need clarity or a deeper understanding of a topic, idea, or any piece of information, utilize the following prompts:\\ \hspace{0.3cm} o Explain [insert specific topic] in simple terms. \\  \hspace{0.3cm} o Explain to me like I'm 11 years old. \\ \hspace{0.3cm} o Explain to me as if  I'm a beginner in [ field   ]. \\ \hspace{0.3cm} o ``Write the [essay/text/paragraph]  using simple English like you're explaining something to a 5-year-old.''\end{tabular}       &
  5  \\
  
 &  \begin{tabular}[c]{@{}l@{}}
  \\ Add to your prompt the following phrase ``Ensure that your answer is unbiased and avoids relying on stereotypes.''  \end{tabular}   &
  13 \\
  
 & \begin{tabular}[c]{@{}l@{}}
  \\ To write any text intended to be similar to a provided sample, include specific instructions:\\ \hspace{0.3cm} o ``Use the same language based on the provided paragraph [/title/text/essay/answer].'' \end{tabular}  &
  26 \\
 & \begin{tabular}{@{}p{10.7cm}@{}}
  \\ When you want to initiate or continue a text using specific words, phrases, or sentences, utilize the provided prompt structure: \\ \hspace{0.3cm} o I’m providing you with the beginning [song lyrics/story/paragraph/essay...]:
[Insert lyrics/words/sentence]. \\ \hspace{0.45cm}  Finish it based on the words provided. Keep the flow consistent. \end{tabular} &
  24 \\
 &  \begin{tabular}{@{}p{10.7cm}@{}}
  \\
  Clearly state the model's requirements that the model must follow in order to produce content, in form of the keywords, regulations, hint, or instructions. \end{tabular}&
  25 \\
 &   \begin{tabular}{@{}p{10.7cm}@{}}
  \\
  To inquire about a specific topic or idea and test your understanding g, you can use the following phrase [16]:\\ \hspace{0.3cm} o ``Teach
me the [Any theorem/topic/rule name] and include a test at the end, and let me know if my answers are correct after I respond, without providing the answers beforehand.'' \end{tabular} &
  15 \\
 &
  \begin{tabular}[c]{@{}l@{}}\\To write an essay/text/paragraph/article or any type of text that should be detailed: \\\hspace{0.3cm} o ``Write a detailed {[}essay/text/paragraph{]} for me on {[}topic{]} in detail by adding all the information necessary.''\end{tabular} &
  21 \\ \hline
\multirow{4}{*}{\begin{tabular}[c]{@{}c@{}}User Interaction \\ and Engagement\end{tabular}} &

 \begin{tabular}{@{}p{10.7cm}@{}}
Allow the model to elicit precise details and requirements from you by asking you questions until he has enough information to provide  the needed output
\\\hspace{0.3cm} o ``From now on, I would like you to ask me questions to ...'' \end{tabular} &
  14 \\
 &
   \begin{tabular}{@{}p{10.7cm}@{}} \\To write an essay /text /paragraph /article or any type of text that should be detailed: ``Write a detailed [essay/text/paragraph] for me on [topic] in detail by adding all the necessary information.''\end{tabular} &
  21 \\ \hline
  
\multirow{16}{*}{\begin{tabular}[c]{@{}c@{}}Content and \\ Language Style\end{tabular}} &
   \begin{tabular}{@{}p{10.7cm}@{}}  To correct/change specific text without changing its style: ``Try to revise every paragraph sent by users. You should only improve  the user's grammar and vocabulary and make sure it sounds natural. You should maintain the original writing style, ensuring that a formal paragraph remains formal.''\end{tabular}  &
  22 \\
 &
 \begin{tabular}[c]{@{}l@{}} \\
  Incorporate the following phrases: ``Your task is'' and ``You MUST.''  \end{tabular}&
  9 \\
 & 
 \begin{tabular}[c]{@{}l@{}} \\
  Incorporate the following phrases: ``You will be penalized.''\end{tabular}  &
  10\\
 & \begin{tabular}[c]{@{}l@{}} \\
  Assign a role to the language model. \end{tabular}  &
  16 \\
 &\begin{tabular}[c]{@{}l@{}} \\
  Use the phrase ``Answer a question given in natural language form'' in your prompts. \end{tabular} &
  11 \\
 & \begin{tabular}{@{}p{10.7cm}@{}} \\
  No need to be polite with LLM so there is no need to add phrases like ``please'', ``if you don't mind'', ``thank you'', ``I would like to'', etc., and get straight to the point.  \end{tabular} &
  1 \\
 &\begin{tabular}[c]{@{}l@{}} \\
  Repeat a specific word or phrase multiple times within a prompt.  \end{tabular}&
  18  \\  
  &\begin{tabular}[c]{@{}l@{}} \\
  Add ``I'm going to tip \$xxx for a better solution!''\end{tabular}&
  6 \\  \hline
\multirow{8}{*}{\begin{tabular}[c]{@{}c@{}}Complex Tasks and \\ Coding Prompts\end{tabular}} 
&
  Break down complex tasks into a sequence of simpler prompts in an interactive conversation.  &
  3 \\
 &
  \begin{tabular}{@{}p{10.7cm}@{}} \\
  When you have a complex coding prompt that may be in different files: \\\hspace{0.3cm} o ``From now and on whenever you
generate code that spans more than one file, generate a [programming language ] script that can be run to
automatically create the specified files or make changes to existing files to insert the generated code. [your
question].''\end{tabular}&
  23 \\
 & \begin{tabular}[c]{@{}l@{}} \\
  Combine Chain-of-thought (Cot) with few-shot prompts.\end{tabular} & 
  19 \\ \hline
\end{tabular}
\caption{Prompt principle categories. }
\label{tab:categories}
\end{table*}